\documentclass{article}

\usepackage[final, nonatbib]{nips_2017}

\usepackage[utf8]{inputenc} 
\usepackage[T1]{fontenc}    
\usepackage{hyperref}       
\usepackage{url}            
\usepackage{booktabs}       
\usepackage{amsfonts}       
\usepackage{nicefrac}       
\usepackage{microtype}      
\usepackage{mathtools}
\usepackage{color}

\title{Neural Discrete Representation Learning}

\author{
  Aaron van den Oord \\
  DeepMind\\
  \texttt{avdnoord@google.com} \\
  \And
  Oriol Vinyals \\
  DeepMind\\
  \texttt{vinyals@google.com} \\
  \And
  Koray Kavukcuoglu \\
  DeepMind\\
  \texttt{korayk@google.com} \\
}

\begin{document}

\maketitle

\begin{abstract}
Learning useful representations without supervision remains a key challenge in machine learning. In this paper, we propose a simple yet powerful generative model that learns such discrete representations. Our model, the Vector Quantised-Variational AutoEncoder (VQ-VAE), differs from VAEs in two key ways: the encoder network outputs discrete, rather than continuous, codes; and the prior is learnt rather than static. In order to learn a discrete latent representation, we incorporate ideas from vector quantisation (VQ). Using the VQ method allows the model to circumvent issues of ``posterior collapse'' -— where the latents are ignored when they are paired with a powerful autoregressive decoder -— typically observed in the VAE framework. Pairing these representations with an autoregressive prior, the model can generate high quality images, videos, and speech as well as doing high quality speaker conversion and unsupervised learning of phonemes, providing further evidence of the utility of the learnt representations.
\end{abstract}
\section{Introduction}\label{sec:intro}

Recent advances in generative modelling of images \cite{van2016conditional, goodfellow2014generative, gregor2016towards, kingma2016improved, dinh2016density}, audio \cite{van2016wavenet, mehri2016samplernn} and videos \cite{kalchbrenner2016video, finn2016unsupervised} have yielded impressive samples and applications \cite{ledig2016photo, isola2016image}. At the same time, challenging tasks such as few-shot learning \cite{santoro2016one}, domain adaptation \cite{hoffman2013efficient}, or reinforcement learning \cite{sutton1998reinforcement} heavily rely on learnt representations from raw data, but the usefulness of generic representations trained in an unsupervised fashion is still far from being the dominant approach.
 
Maximum likelihood and reconstruction error are two common objectives used to train unsupervised models in the pixel domain, however their usefulness depends on the particular application the features are used in. Our goal is to achieve a model that conserves the important features of the data in its latent space while optimising for maximum likelihood. As the work in \cite{chen2016variational} suggests, the best generative models (as measured by log-likelihood) will be those without latents but a powerful decoder (such as PixelCNN). However, in this paper, we argue for learning discrete and useful latent variables, which we demonstrate on a variety of domains.

Learning representations with continuous features have been the focus of many previous work \cite{hinton2006reducing,vincent2010stacked, infogan, denton2016semi} however we concentrate on discrete representations \cite{mnih2014neural,salakhutdinov2009deep,courville2011spike,vimco} which are potentially a more natural fit for many of the modalities we are interested in. Language is inherently discrete, similarly speech is typically represented as a sequence of symbols. Images can often be described concisely by language \cite{vinyals2015show}. Furthermore, discrete representations are a natural fit for complex reasoning,  planning and predictive learning (e.g., if it rains, I will use an umbrella). While using discrete latent variables in deep learning has proven challenging, powerful autoregressive models have been developed for modelling distributions over discrete variables \cite{van2016wavenet}.
 
In our work, we introduce a new family of generative models succesfully combining the variational autoencoder (VAE) framework with discrete  latent representations through a novel parameterisation of the posterior distribution of (discrete) latents given an observation. Our model, which relies on vector quantization (VQ), is simple to train, does not suffer from large variance, and avoids the ``posterior collapse’’ issue which has been problematic with many VAE models that have a powerful decoder, often caused by latents being ignored. Additionally, it is the first discrete latent VAE model that get similar performance as its continuous counterparts, while offering the flexibility of discrete distributions. We term our model the VQ-VAE.

Since VQ-VAE can make effective use of the latent space, it can successfully model important features that usually span many dimensions in data space (for example objects span many pixels in images, phonemes in speech, the message in a text fragment, etc.) as opposed to focusing or spending capacity on noise and imperceptible details which are often local.

Lastly, once a good discrete latent structure of a modality is discovered by the VQ-VAE, we train a powerful prior over these discrete random variables, yielding interesting samples and useful applications.
For instance, when trained on speech we discover the latent structure of language without any supervision or prior knowledge about phonemes or words. Furthermore, we can equip our decoder with the speaker identity, which allows for speaker conversion, i.e., transferring the voice from one speaker to another without changing the contents. We also show promising results on learning long term structure of environments for RL. 

Our contributions can thus be summarised as:
 
\begin{itemize}
\item Introducing the VQ-VAE model, which is simple, uses discrete latents, does not suffer from ``posterior collapse’’ and has no variance issues. 
\item We show that a discrete latent model (VQ-VAE) perform as well as its continuous model counterparts in log-likelihood.
\item When paired with a powerful prior, our samples are coherent and high quality on a wide variety of applications such as speech and video generation.
\item We show evidence of learning language through raw speech, without any supervision, and show applications of unsupervised speaker conversion.
\end{itemize}

\section{Related Work}\label{sec:relwork}

In this work we present a new way of training variational autoencoders \cite{kingma2013auto, rezende2014stochastic} with discrete latent variables \cite{mnih2014neural}. Using discrete variables in deep learning has proven challenging, as suggested by the dominance of continuous latent variables in most of current work -- even when the underlying modality is inherently discrete.

There exist many alternatives for training discrete VAEs. The NVIL \cite{mnih2014neural} estimator use a single-sample objective to optimise the variational lower bound, and uses various variance-reduction techniques to speed up training. VIMCO \cite{vimco} optimises a multi-sample objective \cite{burda2015importance}, which speeds up convergence further by using multiple samples from the inference network. 

Recently a few authors have suggested the use of a new continuous reparemetrisation based on the so-called Concrete \cite{maddison2016concrete} or Gumbel-softmax \cite{jang2016categorical} distribution, which is a continuous distribution and has a temperature constant that can be annealed during training to converge to a discrete distribution in the limit. In the beginning of training the variance of the gradients is low but biased, and towards the end of training the variance becomes high but unbiased. 

None of the above methods, however, close the performance gap of VAEs with continuous latent variables where one can use the Gaussian reparameterisation trick which benefits from much lower variance in the gradients. Furthermore, most of these techniques are typically evaluated on relatively small datasets such as MNIST, and the dimensionality of the latent distributions is small (e.g., below 8). In our work, we use three complex image datasets (CIFAR10, ImageNet, and DeepMind Lab) and a raw speech dataset (VCTK).

Our work also extends the line of research where autoregressive distributions are used in the decoder of VAEs and/or in the prior \cite{gregor2013deep}. This has been done for language modelling with LSTM decoders \cite{bowman2015generating}, and more recently with dilated convolutional decoders \cite{improvedtextvae}. PixelCNNs \cite{oord2016pixel, van2016conditional} are convolutional autoregressive models which have also been used as distribution in the decoder of VAEs \cite{pixelvae, chen2016variational}. 

Finally, our approach also relates to work in image compression with neural networks. Theis et. al. \cite{theis2017lossy} use scalar quantisation to compress activations for lossy image compression before arithmetic encoding. Other authors \cite{agustsson2017soft} propose a method for similar compression model with vector quantisation. The authors propose a continuous relaxation of vector quantisation which is annealed over time to obtain a hard clustering. In their experiments they first train an autoencoder, afterwards vector quantisation is applied to the activations of the encoder, and finally the whole network is fine tuned using the soft-to-hard relaxation with a small learning rate. In our experiments we were unable to train using the soft-to-hard relaxation approach from scratch as the decoder was always able to invert the continuous relaxation during training, so that no actual quantisation took place.

\section{VQ-VAE}\label{sec:method}

Perhaps the work most related to our approach are VAEs.
VAEs consist of the following parts: an encoder network which parameterises a posterior distribution $q(z|x)$ of discrete latent random variables $z$ given the input data $x$, a prior distribution $p(z)$, and a decoder with a distribution $p(x|z)$ over input data.

Typically, the posteriors and priors in VAEs are assumed normally distributed with diagonal covariance, which allows for the Gaussian reparametrisation trick to be used \cite{rezende2014stochastic, kingma2013auto}. Extensions include autoregressive prior and posterior models \cite{gregor2013deep}, normalising flows \cite{rezende2015variational, dinh2016density}, and inverse autoregressive posteriors \cite{kingma2016improved}.

In this work we introduce the VQ-VAE where we use discrete latent variables with a new way of training, inspired by vector quantisation (VQ). The posterior and prior distributions are categorical, and the samples drawn from these distributions index an embedding table. These embeddings are then used as input into the decoder network.

\subsection{Discrete Latent variables}

\begin{figure}
\centering
\includegraphics[width=\textwidth]{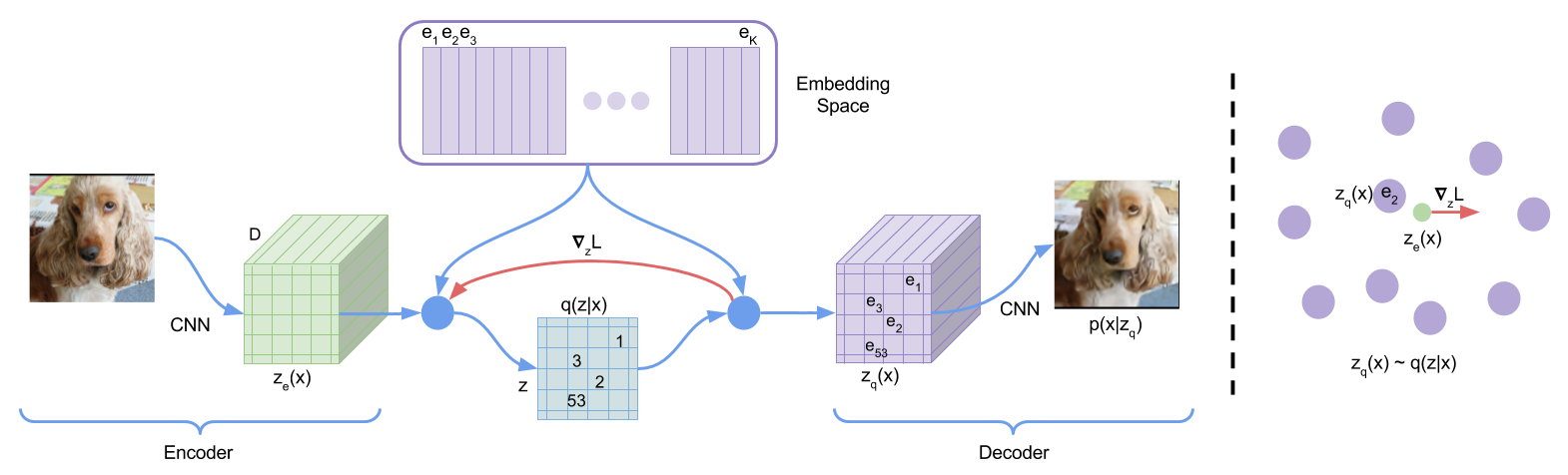}

\caption{Left: A figure describing the VQ-VAE. Right: Visualisation of the embedding space. The output of the encoder $z(x)$ is mapped to the nearest point $e_2$. The gradient $\nabla_z L$ (in red) will push the encoder to change its output, which could alter the configuration in the next forward pass.}
\label{fig:gradient}
\end{figure}

We define a latent embedding space $e \in R^{K \times D}$ where $K$ is the size of the discrete latent space (i.e., a $K$-way categorical), and $D$ is the dimensionality of each latent embedding vector $e_i$. Note that there are $K$ embedding vectors $e_i \in R^D$, $i \in {1, 2, ... , K}$. As shown in Figure~\ref{fig:gradient}, the model takes an input $x$, that is passed through an encoder producing output $z_e(x)$. The discrete latent variables $z$ are then calculated by a nearest neighbour look-up using the shared embedding space $e$ as shown in equation~\ref{eq_assign}. The input to the decoder is the corresponding embedding vector $e_k$ as given in equation~\ref{eq_quantisation}. One can see this forward computation pipeline as a regular autoencoder with a particular non-linearity that maps the latents to $1$-of-K embedding vectors. The complete set of parameters for the model are union of parameters of the encoder, decoder, and the embedding space $e$. For sake of simplicity we use a single random variable $z$ to represent the discrete latent variables in this Section, however for speech, image and videos we actually extract a 1D, 2D and 3D latent feature spaces respectively.

The posterior categorical distribution $q(z|x)$ probabilities are defined as one-hot as follows:

\begin{equation} 
q(z=k|x)=
\begin{cases*}
1 & \text{for } k = \text{argmin}$_j  \|z_e(x) - e_j\|_2$, \\
0 & \text{otherwise} \label{eq_assign}
\end{cases*},
\end{equation} 

where $z_e(x)$ is the output of the encoder network. We view this model as a VAE in which we can bound $\log p(x)$ with the ELBO. Our proposal distribution $q(z=k|x)$ is deterministic, and by defining a simple uniform prior over $z$ we obtain a KL divergence constant and equal to $\log K$.

The representation $z_e(x)$ is passed through the discretisation bottleneck followed by mapping onto the nearest element of embedding $e$ as given in equations~\ref{eq_assign} and~\ref{eq_quantisation}.

\begin{equation}
z_{q}(x) = e_k, \quad \text{where} \quad k = \text{argmin}_j  \|z_e(x) - e_j\|_2
\label{eq_quantisation}
\end{equation}

\subsection{Learning}
\label{section:learning}

Note that there is no real gradient defined for equation~\ref{eq_quantisation}, however we approximate the gradient similar to the straight-through estimator \cite{bengio2013estimating} and just copy gradients from decoder input $z_q(x)$ to encoder output $z_e(x)$. One could also use the subgradient through the quantisation operation, but this simple estimator worked well for the initial experiments in this paper.

During forward computation the nearest embedding $z_{q}(x)$ (equation \ref{eq_quantisation}) is passed to the decoder, and during the backwards pass the gradient $\nabla_z L$ is passed unaltered to the encoder. Since the output representation of the encoder and the input to the decoder share the same $D$ dimensional space, the gradients contain useful information for how the encoder has to change its output to lower the reconstruction loss.

As seen on Figure \ref{fig:gradient} (right), the gradient can push the encoder's output to be discretised differently in the next forward pass, because the assignment in equation \ref{eq_assign} will be different.

Equation \ref{eq_loss} specifies the overall loss function. It is has three components that are used to train different parts of VQ-VAE. The first term is the reconstruction loss (or the data term) which optimizes the decoder and the encoder (through the estimator explained above). Due to the straight-through gradient estimation of mapping from $z_e(x)$ to $z_q(x)$, the embeddings $e_i$ receive no gradients from the reconstruction loss $\log p(z|z_q(x))$. Therefore, in order to learn the embedding space, we use one of the simplest dictionary learning algorithms, Vector Quantisation (VQ). The VQ objective uses the $l_2$ error to move the embedding vectors $e_i$ towards the encoder outputs $z_e(x)$ as shown in the second term of equation~\ref{eq_loss}. Because this loss term is only used for updating the dictionary, one can alternatively also update the dictionary items as function of moving averages of $z_e(x)$ (not used for the experiments in this work). For more details see Appendix \ref{appendix:ema}.

Finally, since the volume of the embedding space is dimensionless, it can grow arbitrarily if the embeddings $e_i$ do not train as fast as the encoder parameters. To make sure the encoder commits to an embedding and its output does not grow, we add a commitment loss, the third term in equation~\ref{eq_loss}. Thus, the total training objective becomes:
\begin{equation}
L = \log p(x|z_q(x)) + \|\text{sg}[z_e(x)] - e\|^2_2 + \beta \|z_e(x) - \text{sg}[e]\|^2_2,
\label{eq_loss}
\end{equation}

where sg stands for the stopgradient operator that is defined as identity at forward computation time and has zero partial derivatives, thus effectively constraining its operand to be a non-updated constant. The decoder optimises the first loss term only, the encoder optimises the first and the last loss terms, and the embeddings are optimised by the middle loss term. We found the resulting algorithm to be quite robust to $\beta$, as the results did not vary for values of $\beta$ ranging from $0.1$ to $2.0$. We use $\beta= 0.25$ in all our experiments, although in general this would depend on the scale of reconstruction loss. Since we assume a uniform prior for $z$, the KL term that usually appears in the ELBO is constant w.r.t. the encoder parameters and can thus be ignored for training.

In our experiments we define $N$ discrete latents (e.g., we use a field of 32 x 32 latents for ImageNet, or 8 x 8 x 10 for CIFAR10). The resulting loss $L$ is identical, except that we get an average over $N$ terms for $k$-means and commitment loss -- one for each latent.

 The log-likelihood of the complete model $\log p(x)$ can be evaluated as follows:
$$
\log p(x) = \log \sum_k p(x|z_k)p(z_k),
$$

Because the decoder $p(x|z)$ is trained with $z = z_q(x)$ from MAP-inference, the decoder should not allocate any probability mass to $p(x|z)$ for $z \neq z_q(x)$ once it has fully converged. Thus, we can write $\log p(x) \approx \log p(x|z_q(x))p(z_q(x))$. We empirically evaluate this approximation in section~\ref{sec:exp}. From Jensen's inequality, we also can write $\log p(x) \geq \log p(x|z_q(x))p(z_q(x))$.

\subsection{Prior}

The prior distribution over the discrete latents $p(z)$ is a categorical distribution, and can be made autoregressive by depending on other $z$ in the feature map. Whilst training the VQ-VAE, the prior is kept constant and uniform. After training, we fit an autoregressive distribution over $z$, $p(z)$, so that we can generate $x$ via ancestral sampling. We use a PixelCNN over the discrete latents for images, and a WaveNet for raw audio. Training the prior and the VQ-VAE jointly, which could strengthen our results, is left as future research.

\section{Experiments}\label{sec:exp}

\subsection{Comparison with continuous variables}

As a first experiment we compare VQ-VAE with normal VAEs (with continuous variables), as well as VIMCO \cite{vimco} with independent Gaussian or categorical priors. We train these models using the same standard VAE architecture on CIFAR10, while varying the latent capacity (number of continuous or discrete latent variables, as well as the dimensionality of the discrete space K). The encoder consists of 2 strided convolutional layers with stride 2 and window size $4\times 4$, followed by two residual $3\times 3$ blocks (implemented as ReLU, 3x3 conv, ReLU, 1x1 conv), all having 256 hidden units. The decoder similarly has two residual $3\times 3$ blocks, followed by two transposed convolutions with stride 2 and window size $4\times 4$. We use the ADAM optimiser \cite{kingma2014adam} with learning rate 2e-4 and evaluate the performance after 250,000 steps with batch-size 128. For VIMCO we use 50 samples in the multi-sample training objective.

The VAE, VQ-VAE and VIMCO models obtain \textbf{4.51} bits/dim, \textbf{4.67} bits/dim and \textbf{5.14} respectively. All reported likelihoods are lower bounds. Our numbers for the continuous VAE are comparable to those reported for a Deep convolutional VAE: \textbf{4.54} bits/dim \cite{gregor2016towards} on this dataset.

Our model is the first among those using discrete latent variables which challenges the performance of continuous VAEs. Thus, we get very good reconstructions like regular VAEs provide, with the compressed representation that symbolic representations provide. A few interesting characteristics, implications and applications of the VQ-VAEs that we train is shown in the next subsections.

\subsection{Images}

Images contain a lot of redundant information as most of the pixels are correlated and noisy, therefore learning models at the pixel level could be wasteful.

In this experiment we show that we can model $x=128\times128\times3$ images by compressing them to a $z=32\times32\times1$ discrete space (with K=512) via a purely deconvolutional $p(x|z)$. So a reduction of $\frac{128\times128\times3\times8}{32\times32\times9}\approx42.6$ in bits. We model images by learning a powerful prior (PixelCNN) over $z$. This allows to not only greatly speed up training and sampling, but also to use the PixelCNNs capacity to capture the global structure instead of the low-level statistics of images.

\begin{figure}[h]
\centering
\includegraphics[width=0.49\textwidth]{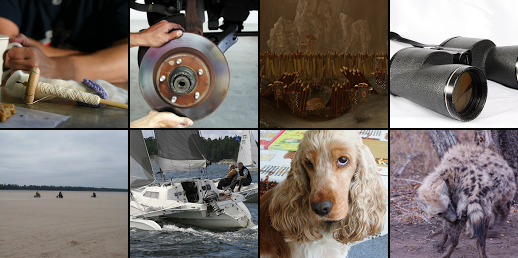}
\includegraphics[width=0.49\textwidth]{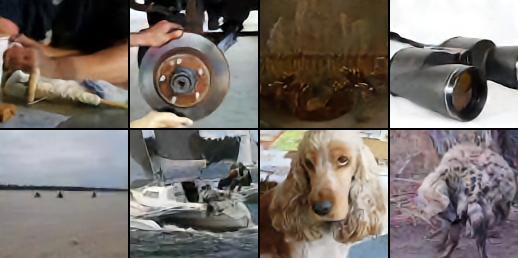}
\caption{Left: ImageNet 128x128x3 images, right: reconstructions from a VQ-VAE with a 32x32x1 latent space, with K=512.}
\label{fig:imnet_recon}
\end{figure}

Reconstructions from the 32x32x1 space with discrete latents are shown in Figure \ref{fig:imnet_recon}. Even considering that we greatly reduce the dimensionality with discrete encoding, the reconstructions look only slightly blurrier than the originals. It would be possible to use a more perceptual loss function than MSE over pixels here (e.g., a GAN \cite{goodfellow2014generative}), but we leave that as future work.

Next, we train a PixelCNN prior on the discretised 32x32x1 latent space. As we only have 1 channel (not 3 as with colours), we only have to use spatial masking in the PixelCNN. The capacity of the PixelCNN we used was similar to those used by the authors of the PixelCNN paper \cite{van2016conditional}. 

\begin{figure}[h]
\centering
\includegraphics[height=0.35\textwidth]{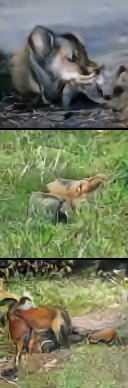}
\includegraphics[height=0.35\textwidth]{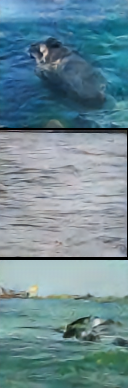}
\includegraphics[height=0.35\textwidth]{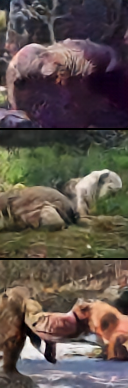}
\includegraphics[height=0.35\textwidth]{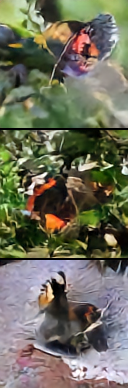}
\includegraphics[height=0.35\textwidth]{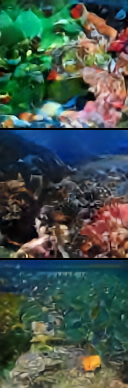}
\includegraphics[height=0.35\textwidth]{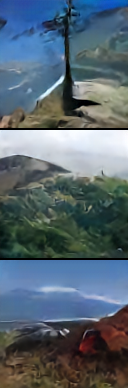}
\includegraphics[height=0.35\textwidth]{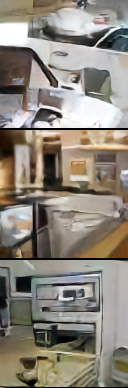}
\includegraphics[height=0.35\textwidth]{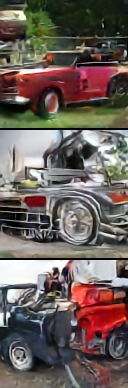}
\caption{Samples (128x128) from a VQ-VAE with a PixelCNN prior trained on ImageNet images. From left to right: kit fox, gray whale, brown bear, admiral (butterfly), coral reef, alp, microwave, pickup.}
\label{fig:imnet_samples}
\end{figure}

Samples drawn from the PixelCNN were mapped to pixel-space with the decoder of the VQ-VAE and can be seen in Figure \ref{fig:imnet_samples}.

\begin{figure}[h]
\centering
\includegraphics[width=\textwidth]{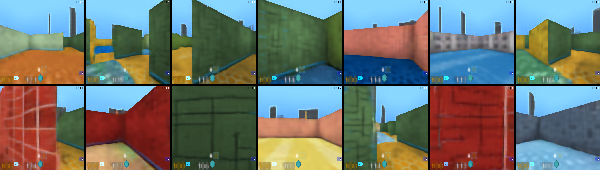}
\caption{Samples (128x128) from a VQ-VAE with a PixelCNN prior trained on frames captured from DeepMind Lab.}
\label{fig:lab_samples}
\end{figure}

We also repeat the same experiment for 84x84x3 frames drawn from the DeepMind Lab environment \cite{beattie2016deepmind}. The reconstructions looked nearly identical to their originals. Samples drawn from the PixelCNN prior trained on the 21x21x1 latent space and decoded to the pixel space using a deconvolutional model decoder can be seen in Figure \ref{fig:lab_samples}.

\begin{figure}[h]
\centering
\includegraphics[width=\textwidth]{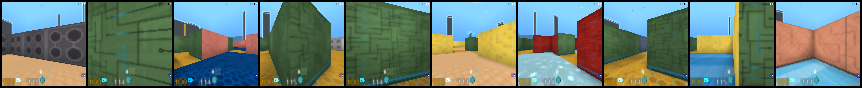}
\includegraphics[width=\textwidth]{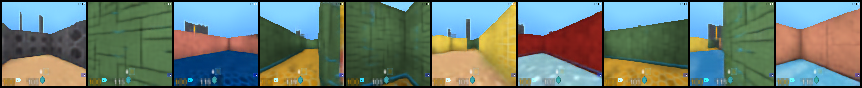}
\caption{Top original images, Bottom: reconstructions from a 2 stage VQ-VAE, with 3 latents to model the whole image (27 bits), and as such the model cannot reconstruct the images perfectly. The reconstructions are generated by sampled from the second PixelCNN prior in the 21x21 latent domain of first VQ-VAE, and then decoded with standard VQ-VAE decoder to 84x84. A lot of the original scene, including textures, room layout and nearby walls remain, but the model does not try to store the pixel values themselves, which means the textures are generated procedurally by the PixelCNN.}
\label{fig:lab_recon_samples}
\end{figure}

Finally, we train a second VQ-VAE with a \emph{PixelCNN decoder} on top of the 21x21x1 latent space from the first VQ-VAE on DM-LAB frames. This setup typically breaks VAEs as they suffer from "posterior collapse", i.e., the latents are ignored as the decoder is powerful enough to model $x$ perfectly. Our model however does not suffer from this, and the latents are meaningfully used. We use only three latent variables (each with K=512 and their own embedding space $e$) at the second stage for modelling the whole image and as such the model cannot reconstruct the image perfectly -- which is consequence of compressing the image onto 3 x 9 bits, i.e. less than a float32. Reconstructions sampled from the discretised global code can be seen in Figure \ref{fig:lab_recon_samples}.

\subsection{Audio}

\begin{figure}[h]
\centering
\includegraphics[width=0.32\textwidth]{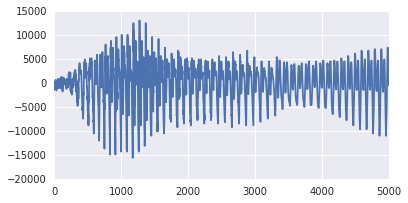}
\includegraphics[width=0.32\textwidth]{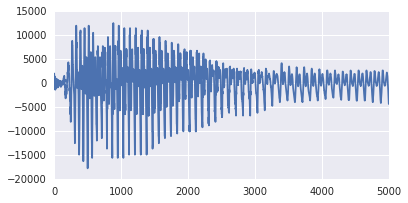}
\includegraphics[width=0.32\textwidth]{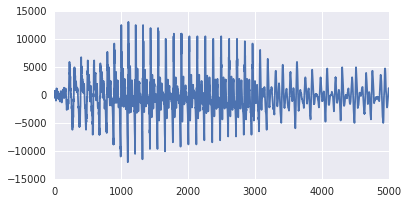}
\caption{Left: original waveform, middle: reconstructed with same speaker-id, right: reconstructed with different speaker-id. The contents of the three waveforms are the same.}
\label{fig:audio}
\end{figure}

In this set of experiments we evaluate the behaviour of discrete latent variables on models of raw audio. In all our audio experiments, we train a VQ-VAE that has a dilated convolutional architecture similar to WaveNet decoder. All samples for this section can be played from the following url: \url{https://avdnoord.github.io/homepage/vqvae/}.

We first consider the VCTK dataset, which has speech recordings of 109 different speakers \cite{yamagishienglish}. We train a VQ-VAE where the encoder has 6 strided convolutions with stride 2 and window-size 4. This yields a latent space 64x smaller than the original waveform. The latents consist of one feature map and the discrete space is 512-dimensional. The decoder is conditioned on both the latents and a one-hot embedding for the speaker. 

First, we ran an experiment to show that VQ-VAE can extract a latent space that only conserves long-term relevant information. After training the model, given an audio example, we can encode it to the discrete latent representation, and reconstruct by sampling from the decoder. Because the dimensionality of the discrete representation is 64 times smaller, the original sample cannot be perfectly reconstructed sample by sample. As it can be heard from the provided samples, and as shown in Figure \ref{fig:action_lab}, the reconstruction has the same content (same text contents), but the waveform is quite different and prosody in the voice is altered. This means that the VQ-VAE has, without any form of linguistic supervision, learned a high-level abstract space that is invariant to low-level features and only encodes the content of the speech. This experiment confirms our observations from before that important features are often those that span many dimensions in the input data space (in this case phoneme and other high-level content in waveform).

We have then analysed the unconditional samples from the model to understand its capabilities. Given the compact and abstract latent representation extracted from the audio, we trained the prior on top of this representation to model the long-term dependencies in the data. For this task we have used a larger dataset of 460 speakers \cite{panayotov2015librispeech} and trained a VQ-VAE model where the resolution of discrete space is 128 times smaller. Next we trained the prior as usual on top of this representation on chunks of 40960 timesteps (2.56 seconds), which yields 320 latent timesteps. While samples drawn from even the best speech models like the original WaveNet \cite{van2016wavenet} sound like babbling , samples from VQ-VAE contain clear words and part-sentences (see samples linked above). We conclude that VQ-VAE was able to model a rudimentary phoneme-level language model in a completely unsupervised fashion from raw audio waveforms.

Next, we attempted the speaker conversion where the latents are extracted from one speaker and then reconstructed through the decoder using a separate speaker id. As can be heard from the samples, the synthesised speech has the same content as the original sample, but with the voice from the second speaker. This experiment again demonstrates that the encoded representation has factored out speaker-specific information: the embeddings not only have the same meaning regardless of details in the waveform, but also across different voice-characteristics.

Finally, in an attempt to better understand the content of the discrete codes we have compared the latents one-to-one with the ground-truth phoneme-sequence (which was not used any way to train the VQ-VAE). With a 128-dimensional discrete space that runs at $25$ Hz (encoder downsampling factor of $640$), we mapped every of the 128 possible latent values to one of the 41 possible phoneme values\footnote{Note that the encoder/decoder pairs could make the meaning of every discrete latent depend on previous latents in the sequence, e.g.. bi/tri-grams (and thus achieve a higher compression) which means a more advanced mapping to phonemes would results in higher accuracy.} (by taking the conditionally most likely phoneme). The accuracy of this 41-way classification was $49.3 \%$, while a random latent space would result in an accuracy of $7.2 \%$ (prior most likely phoneme). It is clear that these discrete latent codes obtained in a fully unsupervised way are high-level speech descriptors that are closely related to phonemes.

\subsection{Video}

\begin{figure}[h]
\centering
\includegraphics[width=\textwidth]{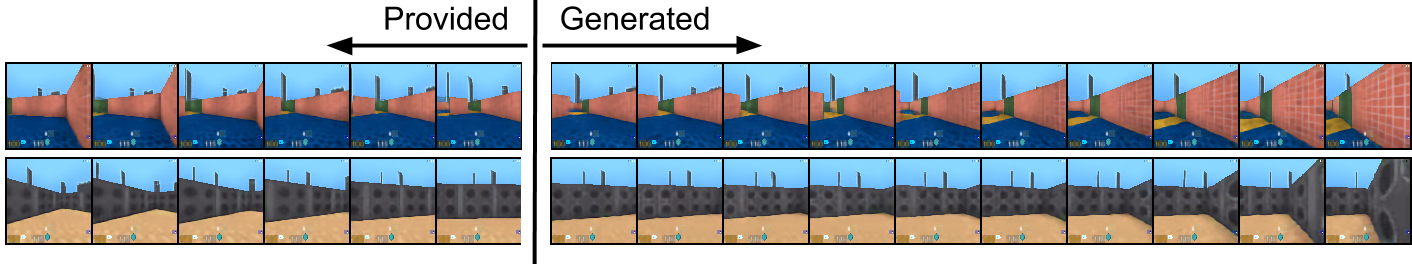}
\caption{First 6 frames are provided to the model, following frames are generated conditioned on an action. Top: repeated action "move forward", bottom: repeated action "move right".}
\label{fig:action_lab}
\end{figure}

For our final experiment we have used the DeepMind Lab \cite{beattie2016deepmind} environment to train a generative model conditioned on a given action sequence. In Figure~\ref{fig:action_lab} we show the initial $6$ frames that are input to the model followed by $10$ frames that are sampled from VQ-VAE with all actions set to \emph{forward} (top row) and \emph{right} (bottom row). Generation of the video sequence with the VQ-VAE model is done purely in the latent space, $z_t$ without the need to generate the actual images themselves. Each image in the sequence $x_t$ is then created by mapping the latents with a deterministic decoder to the pixel space after all the latents are generated using only the prior model $p(z_1, \ldots, z_T)$. Therefore, VQ-VAE can be used to imagine long sequences purely in latent space without resorting to pixel space. It can be seen that the model has learnt to successfully generate a sequence of frames conditioned on given action without any degradation in the visual quality whilst keeping the local geometry correct. For completeness, we trained a model without actions and obtained similar results, not shown due to space constraints.

\section{Conclusion}

In this work we have introduced VQ-VAE, a new family of models that combine VAEs with vector quantisation to obtain a discrete latent representation. We have shown that VQ-VAEs are capable of modelling very long term dependencies through their compressed discrete latent space which we have demonstrated by generating $128\times128$ colour images, sampling action conditional video sequences and finally using audio where even an unconditional model can generate surprisingly meaningful chunks of speech and doing speaker conversion. All these experiments demonstrated that the discrete latent space learnt by VQ-VAEs capture important features of the data in a completely unsupervised manner. Moreover, VQ-VAEs achieve likelihoods that are almost as good as their continuous latent variable counterparts on CIFAR10 data. We believe that this is the first discrete latent variable model that can successfully model long range sequences and fully unsupervisedly learn high-level speech descriptors  that are closely related to phonemes.

\small{
  \bibliography{main}
  \bibliographystyle{plain}
}

\newpage

\appendix 
\section{Appendix}

\subsection{VQ-VAE dictionary updates with Exponential Moving Averages}
\label{appendix:ema}

As mentioned in Section \ref{section:learning}, one can also use exponential moving averages (EMA) to update the dictionary items instead of the loss term from Equation \ref{eq_loss}:
\begin{equation}
\|\text{sg}[z_e(x)] - e\|^2_2.
\label{loss_dict}
\end{equation}

Let $\{z_{i, 1}, z_{i, 2}, \dots, z_{i, n_i}\}$ be the set of $n_i$ outputs from the encoder that are closest to dictionary item $e_i$, so that we can write the loss as:
\begin{equation}
\sum_j^{n_i} \|z_{i, j} - e_i\|^2_2.
\end{equation}
The optimal value for $e_i$ has a closed form solution, which is simply the average of elements in the set:
$$
e_i = \frac{1}{n_i}\sum_j^{n_i} z_{i,j}.
$$
This update is typically used in algorithms such as K-Means.

However, we cannot use this update directly when working with minibatches. Instead we can use exponential moving averages as an online version of this update:
\begin{align}
N^{(t)}_i &:= N^{(t-1)}_i * \gamma + n^{(t)}_i (1 - \gamma) \\
m^{(t)}_i &:= m^{(t-1)}_i * \gamma + \sum_j z^{(t)}_{i,j} (1 - \gamma) \\
e^{(t)}_i &:= \frac{m^{(t)}_i}{N^{(t)}_i}, \label{ema}
\end{align}
with $\gamma$ a value between 0 and 1. We found $\gamma=0.99$ to work well in practice.

\end{document}